\documentclass[journal]{IEEEtran}
\usepackage{algorithmic}
\usepackage{algorithm}
\usepackage{booktabs}
\usepackage{epstopdf}
\usepackage{graphicx}
\usepackage{subfigure}
\usepackage{amsfonts,amssymb}
\usepackage{graphicx}
\usepackage{color}
\usepackage{pifont}
\usepackage{booktabs} 
\newcommand{\eat}[1]{}

\ifCLASSINFOpdf
\else
\fi

\begin{document}
%
\title{From Deterministic to Generative: Multi-Modal Stochastic RNNs for Video Captioning}
%
%
%

\author{Jingkuan Song,
        Yuyu Guo,
        Lianli Gao,
        Xuelong Li,~\IEEEmembership{IEEE Fellow}
        Alan Hanjalic,~\IEEEmembership{IEEE Fellow}
        Heng Tao Shen
\thanks{Jingkuan Song, Yuyu Guo, Lianli Gao and Heng Tao Shen are with the Center of Future Media, University of Electronic Science and Technology of China. Xuelong Li is with Chinese Academy of Sciences, Xi'an, China. Alan Hanjalic is with Delft University of Technology, Netherlands.}
\thanks{Manuscript received July 25, 2017.}}

%
%

\markboth{Journal of \LaTeX\ Class Files,~Vol.~14, No.~8, August~2015}%
{Shell \MakeLowercase{\textit{et al.}}: Bare Demo of IEEEtran.cls for IEEE Journals}
%



\maketitle

\begin{abstract}
\label{abs}Video captioning in essential is a complex natural process, which is affected by various uncertainties stemming from video content, subjective judgment, etc. 
In this paper we build on the recent progress in using encoder-decoder framework for video captioning and address what we find to be a critical deficiency of the existing methods, that most of the decoders propagate deterministic hidden states.
Such complex uncertainty cannot be modeled efficiently by the deterministic models. In this paper, we propose a generative approach, referred to as multi-modal stochastic RNNs networks (MS-RNN), which models the uncertainty observed in the data using latent stochastic variables.
Therefore, MS-RNN can improve the performance of video captioning, and generate multiple sentences to describe a video considering different random factors. 
Specifically, a multi-modal LSTM (M-LSTM) is first proposed to interact with both visual and textual features to capture a high-level representation. Then, a backward stochastic LSTM (S-LSTM) is proposed to support uncertainty propagation by introducing latent variables. 
Experimental results on the challenging datasets MSVD and MSR-VTT show that our proposed MS-RNN approach outperforms the state-of-the-art video captioning benchmarks.
\end{abstract}


%
\IEEEpeerreviewmaketitle

\section{Introduction}
%
%
%
%
\label{sec.intro}

With the explosive growth of online videos over the past decade, video captioning has become a hot research topic. In a nutshell, video captioning is the problem of translating a video into meaningful textual sentences describing its visual content. As such, solving this problem has the potential to help various applications, from video indexing and search to human-robot interaction. 

Building on the pioneering work of Kojima et al. \cite{CAP:Kojima2002Natural}, a series of studies have been conducted to come up with a first generation of video captioning systems \cite{Tra:Lee2008SAVE,Tra:Khan2011Human,Tra:Hanckmann2012Automated}. Recently, however, the development of these systems has more and more relied on deep neuronal networks (DNN) that have been proven effective in both computer vision 
(e.g., image classification and object detection) and natural language understanding (e.g., machine translation and language modeling), forming two technological pillars of video captioning solutions. In particular, deep Convolutional Neural Networks (CNNs) (e.g., VggNet \cite{Net:VGGSimonyan2014Very} and ResNet \cite{Net:ResNetHe_2016_CVPR}) have been widely deployed to extract representative visual features, while Recurrent Neural Networks (RNNs) (e.g., Long Short Term Memory (LSTM) \cite{RNNs:LSTMsep} and Gate Recurrent Unit (GRU) \cite{RNNs:GRUJun} ) have been deployed to translate sequential term vectors to natural language sentences. Despite significant conceptual and computational complexity of these DNN-based models, their effectiveness has given rise to the so-called \textit{encoder-decoder} scheme as a popular modern approach for video captioning. In this scheme, typically a CNN is used as an encoder and a RNN as a decoder. This approach has shown better performance than traditional video captioning methods with hand-crafted features.

\begin{figure}
	\centering
	\includegraphics[width=0.99\linewidth]{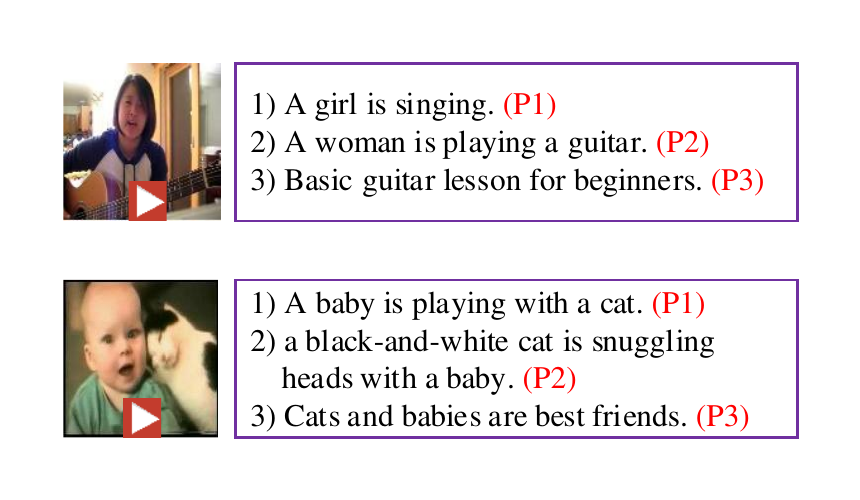}
	\caption{In {real-life} scenario, a video can be described by different sentences because the providers have different intents, experiences and so on. However, if we use deterministic model for video captioning, only one sentence is predicted with the highest probability, which conflicts with the real scenario. By taking different hidden factors (e.g., intention and experience) into consideration, a trained model should be able to output different sentences. P1, P2 and P3 indicates three persons.}
	\label{Fig.demo}
\end{figure}

Recent efforts towards developing and implementing an encoder-decoder scheme for video captioning have mainly focused on solving the following questions: 1) how to help an encode-decoder framework to more efficiently and effectively bridge the gap between video and language \cite{CAPs:seq2seq}? 2) How to facilitate video captioning using semantic information \cite{CAP:EmbedPan2015Jointly}? 3) How to deploy an attention mechanism to help decide what visual information to extract from video \cite{CAP:softattYao2015Describing,CAP:Guo2016Attention}? 4) How to extract attributes/key concepts from sentences to enhance video captioning? \cite{CAP:TM,ImCap:Attri,CAP:BAttri}. Numerous approaches have been proposed to address these questions \cite{CAP:EmbedPan2015Jointly,CAP:HRNNPan2016Hierarchical,CAP:softattYao2015Describing,CAP:Guo2016Attention,CAP:p_RNNYu_2016_CVPR}.

However, the above mentioned approaches have been deterministic without incorporating uncertainties (i.e., both subjective judgment and model uncertainty) into the model calculations at all stages of the modeling. Firstly, in essential, video captioning is a complex process and involves many factors such as video itself, description intents, personal characteristics and experiences, etc. Except for the video content, other factors are inherently random and unpredictable. For example, in Fig.\ref{Fig.demo}, we asked three people to describe two videos separately, and they provided different descriptions for each video. This indicates that video captioning is subjective and uncertain. Secondly, video captioning models are always abstractions of the natural video captioning processes by leaving out some less important components and keeping only relevant and prominent components, thus model uncertainty arises. However, both uncertainties are ignored in previous work.

Therefore, in this paper we are focusing on dealing with the above uncertainties. All our attempts are to ascertain the true nature about video captioning. We propose a novel approach, namely multi-modal stochastic RNN networks (MS-RNN), which model the uncertainty observed in the data using latent stochastic variables. {Our method is inspired by variational auto-encoder (VAE) \cite{VAEs:Bayes}, which uses a set of latent variables to capture the latent information.} Our work makes the following contributions: 
1) We propose an novel end-to-end MS-RNN approach for video captioning. To our knowledge, this is the first approach to video captioning that takes the uncertainty, both subjective judgment and model uncertainty, into consideration. Therefore, for each video, our model can generate multiple sentences to describe it from different aspects.
2) We proposed a multi-modal Long Short-Term Memory (M-LSTM) layer, which incorporates the features from different information sources (i.e., visual and word) into a set of higher-level representation by adjusting the weights on each individual source for improving the video captioning performance. 
3) We develop a novel backward stochastic LSTM (S-LSTM) mechanism to model uncertainty in a latent process through latent variables. With S-LSTM, the uncertainty is expressed in the form of probability distribution of latent variables. The uncertainty can be model into a prior distribution by making use of the consistency between prior distribution and posterior distribution.
4) The proposed model is evaluated on two challenging datasets MSVD and MSR-VTT. The experimental results show that our method achieves superior performance in video captioning. Note that our model only utilizes the appearance features of videos, and no attention mechanism is incorporated.

\vspace{-0.18cm}

\section{Related Work}
\subsection{Recurrent Neural Networks}
Recurrent Neural Networks (RNNs) \cite{RNNs:simpleRNN_2,RNNs:simpleRNN_1} form a directed cycle to connect units. This mechanism allows them to process arbitrary sequential data streams, thus RNNs have been widely used in computational linguistics and achieved great success. Taking language model as an example, RNNs model a sequential data streams (e.g., a sentence) $\textbf{s} = \{s_1,...,s_T\}$ by decomposing the probability distribution over outputs:
\begin{equation}
P(\textbf{s}) = \prod_{t=2}^TP(s_t|s_{<t})P(s_1)
\end{equation}
At each time step, an RNN observes an element and updates its internal states, $h_t = f_{\theta}(h_{t-1},s_t)$, where $f$ is a deterministic non-linear function and $\theta$ indicates a set of parameters. The probability distribution over $s_t$ is parametrized as: $P(s_t|s_{<t}) = P_{\theta}(s_t|h_{t-1})$. The RNN Language Model (RNNLM) \cite{RNNs:RNNLM_Tomas} parametrized the output distribution by applying a softmax function onto the previous hidden state $h_{t-1}$. To learn the model's parameters, RNNLM maximizes the log-likelihood by adopting the gradient descent. 
However, most existing RNNs models propagate deterministic hidden states.
\eat{
However, the above mentioned RNNs are suffering from the "long-term dependencies" problem \cite{RNNs:long-termYoshua,RNNs:GRUJun,RNNs:DGLSTMKai}, as a result LSTM \cite{RNNs:LSTMsep} is proposed for leaning long-term dependencies. Previous studies show that LSTM is capable of modeling data sequences, especially for encoding sentences. Therefore, in this paper, we choose LSTM as our basic component for video captioning.

{ But simple RNNs are absolutely capable of handling "long-term dependencies" \cite{RNNs:long-termYoshua,RNNs:GRUJun,RNNs:DGLSTMKai}. For solving this problem, some variants of RNNs are introduced (i.e., LSTM \cite{RNNs:LSTMsep}, GRU \cite{RNNs:GRUJun} and Depth-Gated LSTM \cite{RNNs:DGLSTMKai}). These variants are explicitly designed several gates inside the internal states to selectively forget or remember information. So some researches always use LSTM instead of normal RNNs to capture long-term temporal information. So in our work, we utilize LSTM to process temporal information.
}
}
{
\subsection{Visual Captioning}

The study of visual captioning problem has been going on for many years.  
In 2002, the video captioning system \cite{CAP:Kojima2002Natural} was proposed for describing human behavior, the method firstly detects visual information (i.e. position of head, direction of head and positions of hands) to find the position where the person is and the gesture what the person does, then selects appropriate predicate, object, etc. with  domain knowledge. Finally, the method applies syntactic rules to generate a whole sentence.
Following this work, a series of studies are conducted to
utilize such technique to enhance different multimedia applications
\cite{Tra:Lee2008SAVE,Tra:Khan2011Human,Tra:Hanckmann2012Automated}.
And there are some works tackle the problem with probabilistic graphical model.
Farhadi \textit{et al.} \cite{Tra:ImCapMRF} introduce the meaning space, which is  represented as triplets of $<$object; action; scene$>$ in the form of a Markov Random Field (MRF), and map the images and sentence to the meaning space to find the relationship of between images and sentences.
Rohrbach \textit{et al.} \cite{Tra:CRF} try to model the relationship between different components of the visual information with a Conditional Random Field (CRF), then tackle the captioning problem as a machine translation problem to generate sentences. 
}

Inspired by the recent advances in image classification using CNN networks (e.g., VggNet \cite{Net:VGGSimonyan2014Very}, GoogLeNet \cite{Net:GNetSzegedy_2015_CVPR} and  ResNet \cite{Net:ResNetHe_2016_CVPR}), and in machine translation utilizing RNN, there have been a few attempts \cite{CAP:EmbedPan2015Jointly,CAP:MP-LSTM,CAP:HRNNPan2016Hierarchical,CAP:p_RNNYu_2016_CVPR,CAP:softattYao2015Describing} to address video caption generation by firstly adopting an efficient CNN network to extract video appearance features, and secondly utilizing a RNN to take video features and the previous predicted words to infer a new word with a softmax \cite{gan2017stylenet,liang2017recurrent,li2017temporal}. In order to further improve the performance, more complex approaches \cite{CAP:softattYao2015Describing,CAP:EmbedPan2015Jointly,CAP:p_RNNYu_2016_CVPR} are proposed from different aspects. Specifically, Yao \textit{et al.} \cite{CAP:softattYao2015Describing} adopted a spatio-temporal convolutional neural network (3-D CNN) for capturing video motion information and a soft attention mechanism to select relevant frame level features for video captioning. Pan \textit{et al.} \cite{CAP:EmbedPan2015Jointly} incorporated the semantic relationship between sentence and visual content for video captioning, while Yu \textit{et al.} \cite{CAP:p_RNNYu_2016_CVPR} proposed a hierarchical framework consisting of a sentence generator to describe a specific short video internal and a paragraph generator to capturing the inter-sentence dependency. However, all of them treat video captioning as a deterministic problem, which can only generate one output, which violate the nature of video captioning. 
By taking different hidden factors (e.g., intention and experience) into consideration, a trained model should be able to output different sentences.
{Note that the model introduced in \cite{ImCap:Diver} also can generate diverse sentences for image captioning, because it use different LSTMs to generate different sentences (the number of LSTMs is equal to the number of different sentences.), so their model no uncertain factors and does not capture the uncertainty in captioning problem.}
  
\subsection{What is Uncertainty}
From the management point of view, uncertainty is the lack of exact knowledge, regardless of what is the cause of this deficiency  \cite{KendallGal2017Uncertainties,LiGal2017Alpha,Gal2016Improving}. Models provide us a solution to clarify our understanding of our knowledge gap, but in real life, understanding the average processes is often not sufficient and it is impossible to predict with certain results \cite{Uusitalo201524}. In general, besides language uncertainty, uncertainty can be classified into six major types \cite{Uusitalo201524,KendallGal2017Uncertainties}: 1) measurement errors resulting from imperfections in measuring devices and observational techniques etc. 2) systematic error, which occurs as the results of bias in the measuring devices or the sampling process. 3) natural variation, which occurs in system that changes, with respect to time, space or other variations, in ways. 4) inherent randomness, which results from a system that is irreducible to a deterministic one. 5) model uncertainty, which mainly arises because the mathematical and computer models that are used for predicting future events or for answering question under specific scenarios. And 6) subjective judgment, which occurs as a result of interpretation of data. Without sufficient data, the experts' judgment will be based on observations and experience. 
All of these uncertainties are hidden factors affecting the results of video captioning, and we propose to model these uncertainties using latent stochastic variables.
{
\subsection{Variational auto-encoder}
As mentioned above, we know that we should find a method to capture the uncertainty in the video captioning problem. But how can we model the uncertainty?
Variational auto-encoder (VAE) \cite{VAEs:Bayes} model gives us a good way to solve this problem. For capturing the variations in the observed variables $\textbf{x}$, the VAE model introduces a set of latent random variables $\textbf{z}$ and rewrites the objective function  $\log P(\textbf{x})$ as follow:
\begin{equation}
\log P(\textbf{x})
\geq E_{Q}[\log P(\textbf{x}|\textbf{z})] -
KL[Q(\textbf{z}|\textbf{x})||P(\textbf{z})] 
:= \mathcal{L}
\end{equation}
where $KL[Q||P] $ is Kullback-Leibler divergence between two distributions $Q$ and $P$, which measures the non-symmetric difference between two probability distributions. And $Q(\textbf{z}|\textbf{x})$ is a approximate posterior distribution, which avoids to solve the intractable true posterior distribution. In \cite{VAEs:Bayes}, the VAE model was used to paint the digits, so it needs to decide not just which number is written, but the angle, the stroke width, and also abstract stylistic properties, so the model uses a set of latent random variables to capture the latent information. Inspired by this, we also use latent variables with a stochastic layer to capture the uncertainty information in the video captioning. Different with painting digits, the video captioning task needs generate different sentences based on the content of the video, so our objective function is a conditional probability, so we use the loss function introduced in conditional variational auto-encoder (CVAE) \cite{CVAE:Kihyuk}, which extend the VAE to dispose conditional probability distribution. And Krishnan \textit{et al.} \cite{VAE:deepkf} compared the different variational models, they guide us to choose a effective variational model. And there are some works extend the VAE model to RNN \cite{VAE:HVRNNIulian,VAE:SRNNMarco,VAE:VRNNJunyoung} for generating speech or music signal. All these works inspire us extend the captioning problem to a uncertainty problem.
\begin{figure*}
	\centering
	\includegraphics[width=1.05\textwidth,height=10.5cm]{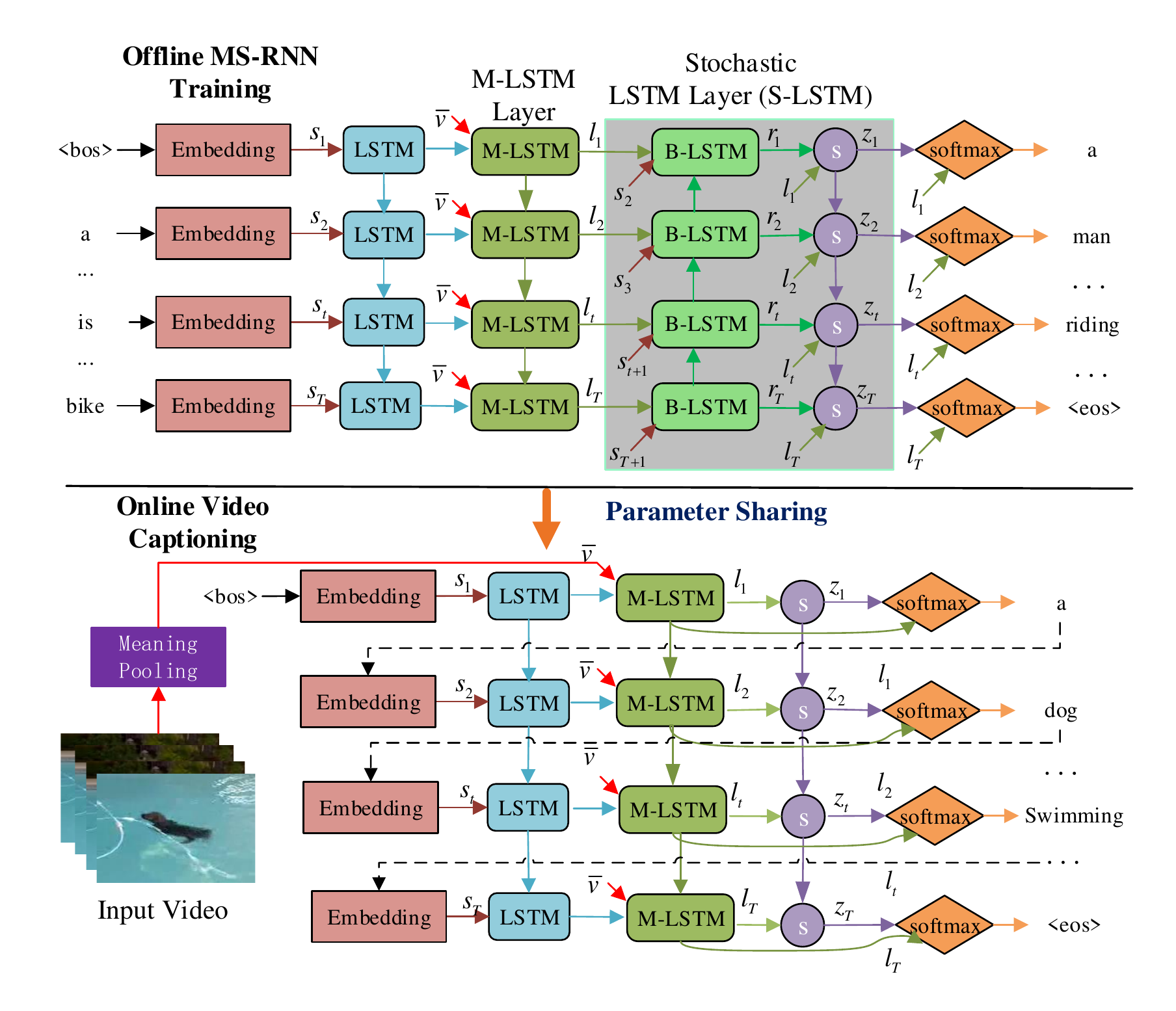}
	\caption{The end-to-end multi-modal RNNs stochastic architecture for video captioning. The S-LSTM is proposed to introduce latent variables to propagate uncertainty. During the training phase, S-LSTM enables the consistency between prior distribution and posterior distribution. Therefore, during the test phase, we only need the learned prior distribution to support video caption generation. { It's a common strategy in VAE model. And we use the B-LSTM to infer the posterior distribution over latent variables, so} the B-LSTM layer is removed during the test phase.}
	\label{Fig.framework}
\end{figure*}
}
\section{The Proposed Approach}
In this section, we introduce our approach for video captioning, and we follow the conventional encoder-decoder framework. 
The encoder is based purely on neural networks to generate video descriptions, and the decoder, named Multimodal Stochastic  Recurrent Neural Networks (MS-RNN) (see Fig.~\ref{Fig.framework}), is our major contribution.
We first introduce the architecture of our proposed network, and then devise the loss function and optimization.


\subsection{Problem Formulation}
Given a video $\textbf{v}$ with $N$ frames, we extract their frame-level features and $\textbf{v}$ can be represented as $\textbf{v} = \{v_1,v_2,...,v_i,...,v_N\}$, where $v_i \in \mathbb{R}^{D_v \times 1}$ and $D_v$ is the dimension of the frame-level features.
For each $\textbf{v}$, we also have a textual sentence $\textbf{a}$ to describe it, and $\textbf{a}$ includes $T$ words which can be represented as $\textbf{a} = \{a_1,a_2,...,a_t,...,a_T\}$. Specifically, $a_t \in \mathbb{R}^{D_a \times 1}$ is one-hot vector where $D_a$ is the dimension of the vocabulary.
Therefore, we have $\textbf{v} \in \mathbb{R}^{D_v \times N}$ and the $\textbf{a} \in \mathbb{R}^{D_a \times T}$. 
Given a video, our model will predict one word at a time until we generate a textual sentence to describe the input video. In detail, in the $t$-th time step, our model utilizes ${\textbf{v}}$ and the previous words $a_{< t}$ to predict a word $a_{t}$ with the maximal probability $P(a_{t}|a_{< t},\textbf{v})$, until we reach the end of the sentence. 
In addition, we set a mark $a_{T+1}=<eos>$ as the end of sentence.

\subsection{Encoder}
The goal of an encoder is to compute feature vectors that are compact and representative and can capture the most related visual information for the decoder.
Specifically, it encodes the input ${\textbf{v}}$ into a continuous representation which may be a variable-sized set $\textbf{v} = \{v_1,v_2,...,v_i,...,v_N\}$.
Thanks to the rapid development of deep convolutional neural networks (CNNs), which have made a great success in large scale image recognition task \cite{Net:ResNetHe_2016_CVPR}, object detection \cite{Nets:faster-rcnn} and visual captioning \cite{CAPs:seq2seq}. High-level features can been extracted from upper or intermediate layers of a deep CNN network. Therefore, a set of well-tested CNN networks, such as the ResNet-152 model \cite{Net:ResNetHe_2016_CVPR} which has achieved the best performance in ImageNet Large Scale Visual Recognition Challenge, can be used as candidate encoders for our framework. With a CNN architecture, we can apply it to each frame to extract representative frame-level features.

For encoding the sentence, because of the sparsity of one-hot vectors $\textbf{a} = \{a_1,a_2,...,a_t,...,a_T\}$, like previous works \cite{CAP:softattYao2015Describing,CAP:EmbedPan2015Jointly}, we process one-hot vector with "embedding" method. We set a parameter matrix $\textbf{U}_{s} \in \mathbb{R}^{D_{s} \times D_{a}}$ to map the one-hot vectors  $\textbf{a}$ to $\textbf{s}$ as follow: 
\begin{equation}
\label{equ.embed}
\textbf{s} = {\textbf{U}_{s} \textbf{a}}
\end{equation}
The $\textbf{s} \in \mathbb{R}^{D_{s} \times T}$ and $\textbf{s} = \{s_1,s_2,...,s_t,...,s_{T}\}$ will be input to the next step. In addition, the end of sentence $a_{T+1}=<eos>$ is mapped to $s_{T+1}$.

\subsection{Decoder with MS-RNN}
The MS-RNN consists of three core components as shown Fig. \ref{Fig.framework}: a basic LSTM layer for extracting word-level features, a multimodal LSTM layer (M-LSTM) for encoding multi-view information (visual and textual features) simultaneously and chronologically, and a backward stochastic LSTM layer (S-LSTM) to adequately introduce latent variables.
\eat{ 
	to compute the prior distribution and posterior distribution.}

\subsubsection{LSTM for Word Features}
In our MS-RNN model, we use a basic LSTM layer to take $\textbf{s} = \{s_1,s_2,...,s_t,...,s_T\}$ as input and output word features $\textbf{s}' = \{s'_{1},s'_{2},...,s'_t,...,s'_{T}\} $ with encoded temporal information.
\begin{equation}
s'_{t} = LSTM(\ s_t, ~s'_{t-1}),~~~~t\in\{1,2,...,T\}
\end{equation}
where $s'_{0}=\textbf{0}$. More specifically, a standard LSTM unit consists of three gates: a ``forget gate'' ($f_t$) that decides what information we are going to throw away from a LSTM unit; an ``input gate'' ($i_t$) that decides what new information we are going to store in the cell state; and an ``output gate'' $o_t$ that controls the extent to which the value in memory is used to compute the output activation of the block. A standard LSTM can be defined as: 
\begin{equation}
\begin{array}{l}
f_t = \sigma(W_{xf} s_t + W_{hf} s'_{t-1} + b_f) \\
i_t = \sigma(W_{xi} s_t + W_{hi} s'_{t-1} + b_i) \\
o_t = \sigma(W_{xo} s_t + W_{ho} s'_{t-1} + b_o) \\ 
g_t = \phi(W_{xg} s_t + W_{hg} s'_{t-1} + b_g) \\
c_t = f_t \odot c_{t-1} + i_t \odot g_{t} \\
s'_t = o_t \odot \phi \left( {{c_t}} \right)
\end{array}
\end{equation}
where $\sigma(\cdot) $ is a sigmoid function, $\phi(\cdot)$ denotes a hyperbolic tangent function, $c_t$ is a cell state vector, $s'_t$ is an output vector, $g_t$ is a sigmoid gate, $W_{*}$ is a set of parameters, $\odot$ denotes the element-wise multiplication, and $b_*$ is a set of bias values.
Then, for each word $s_t$, we extracted its word features as $s'_t$. 

\eat{
{Given a set of inputs $\textbf{s} = \{s_1,s_2,...,s_t,...,s_T\}$, two initialized vectors $s'_{0}$ and $c_0$,} 
}

\subsubsection{Multimodal LSTM Layer}
Next, a M-LSTM layer takes the $\textbf{s}'$ and a video-level feature $\overline{v}$ as inputs to fuse a high-level features $l_t$. 
\begin{equation}
l_t = M\_LSTM(s'_t,\ \overline{v},\ l_{t-1})\ \ \ \ \ t \in \{1,2,...,T\} 
\end{equation}
Here, instead of using advanced but complex temporal or spatial attention mechanism to select a video-level feature, we use the basic mean pooling strategy to obtain one $\overline{v}$:
\begin{equation}
\overline{v} = \frac{1}{N} \sum_{i=1}^{N} v_i,~~~~~~~ v_i \in \textbf{v}
\label{equ_mean_v}
\end{equation}
The motivation is that if our model using the basic way to utilize the visual features can improve the performance of video captioning, the advantages of our MS-RNN are manifest. However, as shown in \cite{CAP:softattYao2015Describing,CAP:Guo2016Attention}, the attention mechanism can further boost the performance of video captioning.

Multimodal LSTM (M-LSTM) is a novel variant of LSTM, and it not only inherits the numerical stability of LSTM but also generates plausible features from multiview sources. We choose LSTM as our basic RNN unit due to the following reasons: 1) it achieved great success in machine translation, speech recognition, image and video caption \cite{ImCap:LearingRNN,CAPs:From_captions,CAPs:seq2seq}; and 2) compared with basic RNN units, it is absolutely capable of handing the ``long-term dependencies'' problem.

Given two modalities $\textbf{s}' = \{s'_{1},s'_{2},...,s'_t,...,s'_{T}\} $ and $\overline{v}$ as the inputs, and two initialized vectors $l_{0}$ and $c_0$, a M-LSTM can be used to fuse them and extract a higher-level feature. A M-LSTM unit can be described as bellow:
{
\begin{equation}
\begin{array}{l}
\label{euq.m-lstm}
i_t = \sigma(W'_{xi} s'_t + W'_{hi} l_{t-1} + W'_{yi} \overline{v}_t + b'_i) \\
f_t = \sigma(W'_{xf} s'_t + W'_{hf} l_{t-1} + W'_{yf} \overline{v}_t + b'_f) \\
o_t = \sigma(W'_{xo} s'_t + W'_{ho} l_{t-1} + W'_{yo} \overline{v}_t + b'_o) \\ 
g_t= \phi  (W'_{xg} s'_t + W'_{hg} l_{t-1} + W'_{yg} \overline{v}_t + b'_g) \\
c_t = f_t \odot c_{t-1} + i_t \odot g_{t} \\
l_t= o_t \odot \phi (c_t)
\end{array}
\end{equation}
}
To obtain an abstract concept from two-modalities, the M-LSTM needs to firstly project $s'_t$ and $\overline{v}$ into a common feature space, then the inside gates can add them together with an activation function. 
\eat{
	Specifically, for parameter initializations of $W_{x*}$ and $W_{y*}$, the random scaled Gaussian initialization has a norm-preserving nature. However, in \cite{Par:ORTHINI}, it shows that the random orthogonal initiation performs better than the random scaled Gaussian initialization, since the random orthogonal initiation is able to achieve depth independent learning times. Therefore, in our M-LSTM we initialize the $W_{x*}$ using the random orthogonal initiations, while  $W_{y*}$ is initialized with the random scaled Gaussian initialization method.
	In such way, we can use $W_{x*}$ and $W_{y*}$ to project $x_t$ and $y_t$ into a common feature space. }
Then, in each time step $t$, we extracted a higher-level feature $l_t$.
\eat{For convenience, we demote the M-LSTM as bellow: 
	\begin{equation}
	h_t,\ c_t = M\_LSTM(x_t,\ y_t,\ h_{t-1}, \ c_{t-1}) \ \ \ \ \ t\in\{1,2,...,T\}
	\end{equation} }

\begin{figure}[b]
	\centering
	\includegraphics[width=0.95\linewidth]{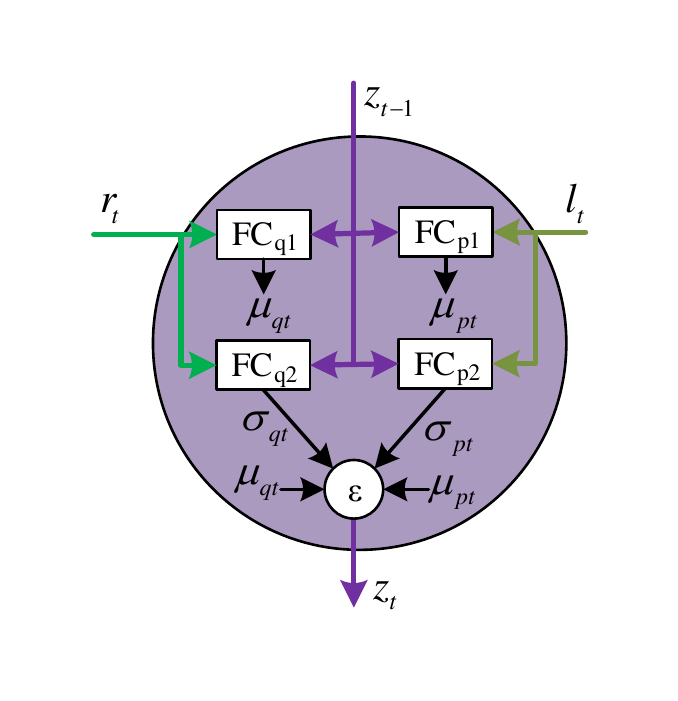}
	\caption{The stochastic cell of the S-LSTM.}
	\label{Fig.srnn_unit}
\end{figure}
\subsubsection{Backward Stochastic LSTM}
\label{sec.BSLSTM}
In this subsection, we introduce our Backward Stochastic LSTM (S-LSTM) to take the output of M-LSTM to approximate the posterior distributions over latent variables defined as $\textbf{z}=\{z_1,z_2,...,z_T\}$, where $z_t \in \mathbb{R}^{D_z}$.
The S-LSTM consists of two units: a backward LSTM unit and a stochastic unit. We define the output of the backward LSTM as $r_t$.

For the backward LSTM unit in time step $t$, its output is defined as:
\begin{equation}
r_t = B\_LSTM(s_{t+1},\ l_t,\ r_{t+1}) \ \ \ \ \ t\in\{1,2,...,T\}
\end{equation}
where $l_t$ is the output of M-LSTM at time step $t$, $s_{t+1}$ is the output of  embedding layer, and  $r_{t+1}$ is initialized to zero vector. The form of $B\_LSTM$ similar with $M\_LSTM$, but it process sequence with backward direction. We can see that the output of backward LSTM in time step $t$ depends on the present input $l_t$, $s_{t+1}$ and future output $r_{t+1}$. This is because in the stochastic units, the posterior distribution of $z_{t}$, which is calculated with Eq.~\ref{eq_kl1}, does not depend on the past outputs and deterministic states, but depend on the present and future ones. 
Therefore, we propose to use the backward LSTM to extract the future information, and incorporate it with a stochastic layer to achieve our goal.

Fig.~\ref{Fig.srnn_unit} demonstrates the stochastic unit structure. To obtain $z_t$, we utilize an ``reparameterization trick'' introduced in \cite{VAEs:Bayes}. This trick randomly samples a set of values $\epsilon_t \in \mathbb{R}^{D_z}$ from a standard Gaussian distribution. Therefore, $\epsilon_t \sim \mathcal{N}(0,1)$. If we assume  $z_t \sim \mathcal{N}(\mu_t,diag(\sigma^2_t))$, we can use $z_t = \mu_t + \sigma_t \odot \epsilon_t$ to calculate $z_t$. Next, we need to solve the problem of how to learn $\mu_t$ and $\sigma_t$ for $z_t$.

In detail, the stochastic unit takes $l_t$, and $z_{t-1}$ as input to approximate $\mu_{p_t}$ and $ \sigma_{p_t} $ by two feed-forward networks (i.e., $FC_{p1}$ and $FC_{p2}$). In addition, each of them contains two fully connected layers.
\begin{equation}
\begin{array}{l}
\mu_{p_t} = FC_{p1}([z_{t-1},l_t])\\
\sigma_{p_t} = \exp \left( 0.5\times {FC_{p2}([z_{t-1},l_t])} \right) 
\end{array}
\end{equation}

$[z_{t-1},l_t]$ is a concatenation operation. In addition, the stochastic unit also takes $r_t$ and $z_{t-1}$ to approximate $\mu_{q_t}$ and $ \sigma_{q_t}$ by two feed-forward networks (i.e., $FC_{q1}$ and $FC_{q2}$):

\begin{equation}
\begin{array}{l}
\mu_{q_t} = FC_{q1}([z_{t-1},r_t])\\
\sigma_{q_t} = \exp \left(0.5 \times {FC_{q2}([z_{t-1},r_t])} \right) 
\end{array}
\end{equation}

{For training, we set $z_t = \mu_{q_t} + \mu_{p_t} +\sigma_{q_t} \odot \epsilon_t$, this method, introduced in \cite{VAE:SRNNMarco}, can improve the posterior approximation by using the prior mean, while for testing we set  $z_t = \mu_{p_t} + \sigma_{p_t} \odot \epsilon_t$, and we set the $z_0$ as zero vector at the beginning.
}
To output a symbol $a_{t}$, a probability distribution over a set of possible words is obtained using $\mathbf{U}_{p}$ and $z_t$:
{
\begin{equation}
\label{softmax}
P(a_{t+1}|z_t,l_t) = softmax \left( \mathbf{U}_{p} [z_t,l_t] + \mathbf{b}  \right)
\end{equation}}where $\mathbf{U}_{p}$ and $\mathbf{b}$ are parameters to be learned. Next, we can interpret the output of the softmax layer $P(a_{t+1}|z_t,l_t)$ as a probability distribution over words.

\subsection{Loss Function}
Based on the variational inference and conditional variational autoencoder (CVAE) proposed in \cite{CVAE:Kihyuk}, we define the following loss function:
\begin{equation}
\log P(\textbf{a}|\textbf{l})
\eat{
	\eat{$A = \{u_t,\textbf{v}\} $ }
	= \log \sum_{t=1}^{T}P(\textbf{s},z_{t}|A)\\
	= \log \sum_{t=1}^{T}\frac{P(\textbf{s},z_{t},A)}{P(z_{t},A)}
	\frac{P(z_{t},A)}{P(A)} \\
	= \log \sum_{t=1}^{T}P(\textbf{s}|z_{t},A)P(z_{t}|A) \\
	\geq \log \sum_{t=1}^{T}Q(z_{t}|\textbf{s},A)\frac{Q(z_{t}|\textbf{s},A)}{Q(z_{t}|\textbf{s},A)}P(\textbf{s}|z_{t},A)P(z_{t}|A) \\
	\geq \sum_{t=1}^{T} \log Q(z_{t}|\textbf{s},A)\frac{Q(z_{t}|\textbf{s},A)}{Q(z_{t}|\textbf{s},A)}P(\textbf{s}|z_{t},A)P(z_{t}|A) \\
	\geq E_{Q}[\log P(\textbf{s}|\textbf{z},A] - KL[Q(\textbf{z}|\textbf{s},A)||P(\textbf{z}|A)] \\ 
}
\geq E_{Q}[\log P(\textbf{a}|\textbf{z},\textbf{\emph{l}})] -
KL[Q(\textbf{z}|\textbf{a},\textbf{\emph{l}})||P(\textbf{z}|\textbf{\emph{l}})] 
:= \mathcal{L}
\label{lossfunction}
\end{equation}
where $\mathcal{L}$ is the evidence lower bound (ELBO) of the log likelihood. The distribution $Q(\textbf{z}|\textbf{a},\textbf{\emph{l}})$ is an approximate posterior distribution , which aims to approximate the intractable true posterior distribution.  For the first term $E_{Q}[\log P(\textbf{a}|\textbf{z},\textbf{\emph{l}})]$, which is an expected log likelihood under $Q(\textbf{z}|\textbf{a},\textbf{\emph{l}})$. This term is written as:
\begin{equation}
\begin{array}{l}
	E_{Q}[\log P(\textbf{a}|\textbf{z},\textbf{\emph{l}})] \\
	= E_{Q}[\sum_{t=1}^{T} \log P(a_{t+1}|z_t,l_t)] \\
	= \sum_{t=1}^{T}\log P(a_{t+1}|z_t,l_t)
\end{array}
\end{equation}
Here, we process the concatenation vector $[z_t,l_t]$ with a softmax layer, mentioned by Eq.\ref{softmax}, to approximate $P(a_{t+1}|z_t,l_t)$.

The second term $KL[Q(\textbf{z}|\textbf{a},\textbf{\emph{l}})||P(\textbf{z}|\textbf{\emph{l}})]$ , namely KL term, is the Kullback-Leibler divergence, which measures the non-symmetric difference between two probability distributions (i.e., $Q(\textbf{z}|\textbf{a},\textbf{\emph{l}})$ and $P(\textbf{z}|\textbf{\emph{l}})$). And in our work, we choose the variational model introduced in \cite{VAE:deepkf} to factorize the  posterior distribution. The posterior and prior distributions are factorized as below:

\begin{equation}
Q(\textbf{z}|\textbf{a},\textbf{\emph{l}}) = \prod_{t=1}^{T} Q(z_t|z_{t-1},a_{>t},l_{\geq t}) Q(z_0|a_{>0},l_{\geq 0})  
\label{eq_kl1}
\end{equation}
\eat{
	\begin{equation}
	Q(\textbf{z}|\textbf{s},\textbf{\emph{l}}) = \prod_{t=1}^{T} Q(z_t|z_{t-1},\textbf{s},\textbf{\emph{l}}) Q(z_0|\textbf{s},\textbf{\emph{l}})  
	\label{eq_kl1_1}
	\end{equation}
}
\begin{equation} 
P(\textbf{z}|\textbf{\emph{l}}) = \prod_{t=1}^{T} P(z_t|z_{t-1},l_t)P(z_0|l_0)
\label{eq_kl2}
\end{equation}
For  approximating the $Q(z_t|z_{t-1},a_{>t},l_{\geq t})$ and $ P(z_t|z_{t-1},l_t)$, we firstly use a  backward LSTM layer to encode $s_{t+1}$ ( we have encoded $a_{t+1}$ to $s_{t+1}$ mentioned in Eq.\ref{equ.embed}) and $l_t$ to $r_t$, then utilize the method, mentioned in Sec. \ref{sec.BSLSTM}, to approximate the means and the variances of  $Q(z_t|z_{t-1},a_{>t},l_{\geq t})$ and $ P(z_t|z_{t-1},l_t)$. 
So we can use the following function to calculate the Kullback-Leibler divergence at the $t$-th time step:

\begin{equation}
\begin{array}{l}
KL[Q_{t}||P_{t}] =   \sum_{i=1}^{D_{z}}\log Q(z_{t_i}|z_{t-1},a_{>t},l_{\geq t})\frac{P(z_{t_i}|z_{t-1},l_t)}{Q(z_{t_i}|z_{t-1},a_{>t},l_{\geq t})}  \\
=  \sum_{i=1}^{D_{z}}\log \frac{\sigma_{p_{t_i}}}{\sigma_{q_{t_i}}} + \frac{\sigma^2_{q_{t_i}} + (\mu_{q_{t_i}} - \mu_{p_{t_i}})^2}{2\sigma^2_{p_{t_i}}} - \frac{1}{2}
\end{array}
\end{equation}

For the whole sentence generation, we calculate the global Kullback-Leibler divergence $KL[Q(\textbf{z}|\textbf{a},\textbf{\emph{l}})||P(\textbf{z}|\textbf{\emph{l}})]$ by:

\begin{equation}
KL[Q(\textbf{z}|\textbf{a},\textbf{\emph{l}})||P(\textbf{z}|\textbf{\emph{l}})] =
\sum_{t=1}^{T} KL[Q_{t}||P_{t}]
\end{equation}

In this paper, we maximize the above proposed loss function to learn all the parameters. More specifically, we use backpropagation through time (BPTT) algorithm to compute the gradients and conduct the optimization with adadelta \cite{OPT:ADADELTA}.

\section{Experiment}
We evaluate our model on two standard video captioning benchmark datasets: the widely used Microsoft Video Description (MSVD) \cite{Dataset:msvd} and the large-scale MSR Video-to-Text (MSR-VTT) \cite{Dataset:msr-vtt}.

\noindent\textbf{MSVD}: This dataset consists of $1,970$ short video clips collected from YouTube, with an average length of about 9s. In addition, this dataset contains about 80,000 clip-description pairs labeled by Amazon Mechanical Turkers (AMT). In other words, each clip has multiple sentence descriptions. In total, all the descriptions contain nearly $16,000$ unique vocabularies. Following previous work  \cite{CAP:EmbedPan2015Jointly,CAP:HRNNPan2016Hierarchical,CAP:p_RNNYu_2016_CVPR}, we split this dataset into a training, a validation and a testing dataset with $1200$, $100$ and $670$ video clips, respectively. 

\noindent\textbf{MSR-VTT}:
This dataset was proposed by Xu \textit{et al.} \cite{Dataset:msr-vtt} in 2016. They aim to provide a new large-scale video benchmark for supporting video understanding, especially for the task of translating videos to text. In total, this dataset contains $10$K web video clips and $200$K clip-sentence pairs in total. Each clip is annotated with 20 natural natural sentences by $1,327$ AMT workers. This dataset is collected from a commercial video search engining and so far it covers the most comprehensive categories and diverse visual content, representing the largest dataset in terms of sentences and vocabularies. We run our experiments on their updated version with sentence quality control. This dataset is divided into three subsets: $65$\% for training, $5$\% for validating and $30$\% for testing. 
  
 \begin{figure*}[t]
 	\centering
 	\includegraphics[width=1\textwidth]{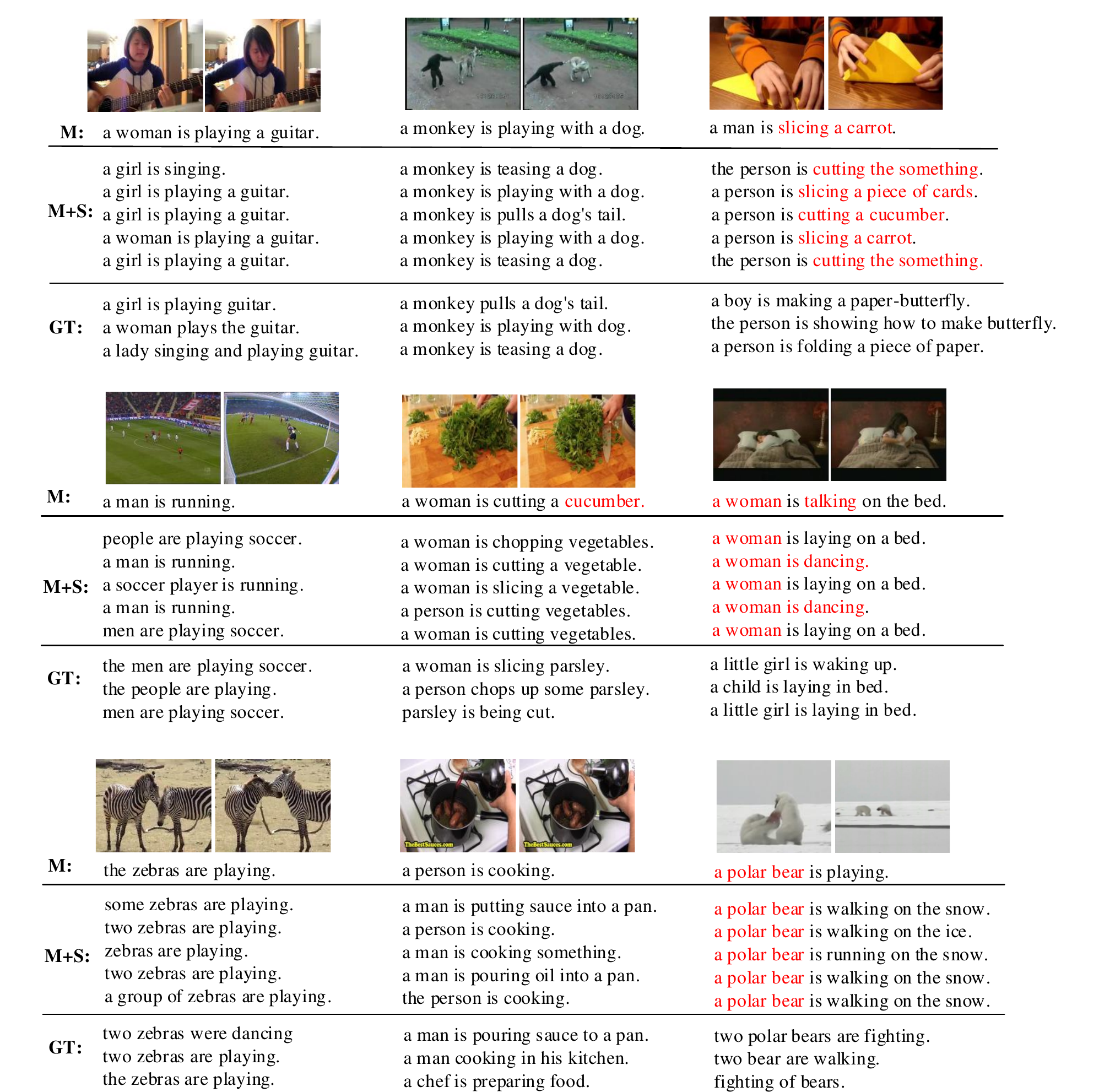}
 	\caption{Demonstration of our results, which are generated by repeatedly inputing each video five times into our trained model on the MSVD dataset. Our model is able to generate different captions based on the different hidden stochastic variables.}
 	\label{five-rs}
 \end{figure*}

\subsection{Evaluation Metrics}
To evaluate the performance of our model, we utilize the following four evaluation metrics: BLUE \cite{EVA:BLUE}, METEOR \cite{EVA:METEOR}, CIDEr \cite{EVA:CIDEr} and  ROUGE-L \cite{EVA:ROUGE}. In addition, Microsoft COCO evaluation server \cite{EVA:MCOCO} has implemented these metrics, so we directly call such evaluation functions to test the performance of video captioning.
   
\subsection{Experimental Settings}
\textbf{Video Appearance Feature Extraction.}
The experimental results obtained by Xu \textit{et al.} \cite{Dataset:msr-vtt} show that applying different pooling methods (i.e., single frame, meaning pooling and soft-attention) obtains different performance. Both mean pooling and soft-attention perform significantly better than single frame. The soft-attention performs slightly better than mean pooling with 0.6\% BULE@4 and 0.6\% METEOR increases, but it involves more operations. Therefore, we apply a mean pooling to a set of frame level features to generate a representative video-level feature. In addition, we follow previous work \cite{CAP:softattYao2015Describing} to uniformly sample $K=28$ frames from each clip for controlling video frames duplication. Deep convolutional neural networks (CNNs) achieved a great success in image feature extraction. Therefore, in this paper we respectively use the ResNet-152 \cite{Net:ResNetHe_2016_CVPR} and GoogLeNet \cite{Net:GNetSzegedy_2015_CVPR}, the two state-of-the-art CNNs, to extract video frame level features to analyze our model. The results show that ResNet features perform better (see Tab\ref{tab-vf}).

\textbf{Sentence Preprocessing.}
For MSVD dataset, we tokenize it by firstly converting all words to lowercases and secondly utilizing the WordPunct function from NLTK toolbox to tokenize sentences and remove punctuations. As a result, we obtained a vocabulary with $13,010$ words from the MSVD training dataset. For the MSR-VTT dataset, after tokenization we obtain a $23,662$ size vocabulary from its training dataset. For each dataset, we use the one-hot vector ($1$-of-$N$ encoding, where $N$ is the vocabulary size) to represent each word.

\textbf{Training Details.} For dealing with sentences with arbitrary size, we add a begin-of-sentence tag $<$bos$>$ to start each sentence, and an end-of-sentence tag $<$eos$>$ to end each sentence. During training, we maximize the loss function by taking the video and its corresponding groundtruth sentence label as the inputs.

In addition, in our experiments, with an initial learning rate $10^{-4}$ to avoid the gradient explosion, we set the beam search size as $5$. Empirically, we set all the M-LSTM unit sizes as 512, all the B-LSTM unit sizes as 512, the dimension of latent variables as 256, and the word embedding size as 512. Our objective function Eq.\ref{lossfunction} is optimized over the whole training video sentence pairs with mini-batch 64 in size of MSVD and 256 in size of MSR-VTT. We stop training our model until 500 epochs are reached or until the evaluation metric does not improve on the validation set at the patience of 20. In addition, we multiply the KL term by a scalar, which starts at 0.01 and linearly increases to 1 over the first 20 epochs.

\textbf{Testing Details.} During testing, our model takes the video and a begin-of-sentence tag $<$bos$>$ as inputs to generate sentences to describe the input video. After the parameter are learned, we perform the generation with Beam Search \cite{method:BeamSearch}.  

In addition, our model incorporates latent variables for ascertaining the true nature about video caption and has potential to describe video from different aspects. Thus, we have repeatedly input the test videos into our trained model five times. Each time we obtain a performance showing in Tab.\ref{Tab.mean}. Finally, we obtain an average performance. Moreover, Fig.\ref{five-rs} shows some output examples.

\begin{table}[t]
	\centering
	\caption{Performances of our MS model obtained by repeatedly input test videos into our model five times.}
		\label{Tab.mean}
	\begin{tabular}{c||c|c|c|c|c|c|c}
		\hline
		Time&B@1&B@2&B@3&B@4&M&C&RL\\
		\hline
		1& \textbf{83.2}&\textbf{72.8}&63.5&53.4&33.9&73.7&\textbf{70.4}\\
		2&82.7&72.6&\textbf{63.7}&53.6&\textbf{34.0}&75.2&70.3\\
		3&82.4&72.1&62.9&52.8&33.6&74.7&69.8\\
		4&83.0&72.8&63.6&\textbf{53.8}&\textbf{34.0}&76.6&\textbf{70.4}\\
		5&83.1&72.7&63.5&53.1&33.6&73.9&70.0\\
		\hline
		mean&82.9&72.6&63.5&53.3&33.8&74.8&70.2\\
		\hline
	\end{tabular}

\end{table}

\subsection{Results on MSVD Dataset}
In this paper, we propose to utilize probability distribution of latent variables to depict uncertainty, thus for each time our model may generate different descriptions. In this subsection, we run the testing five times and report the results in Tab.\ref{Tab.mean}. The performance of each testing is quite stable and reasonable. By checking the generated sentences (see Fig.\ref{five-rs}), we can see that our model can describe a video from various aspects, likely in real life, human provide various sentences to describe one video to fit their intents.
\subsection{Component Analysis} 
In this paper, we design two core components: a M-LSTM layer and a S-LSTM layer, which affect the performance of our algorithm. In this subsection, we study their performance variance with the following two settings:
\begin{itemize}
	\item Only using M-LSTM for video captioning (M).
	\item Incorporating M-LSTM and S-LSTM for video captioning (M+S). 
\end{itemize}

In this sub-experiment, we firstly conduct the experiments on the MSVD dataset and use ResNet to extract frame features. 

Tab.\ref{tab-setting} lists the results, which demonstrate that our MS-RNN model with both M-LSTM and S-LSTM outperforms M-LSTM only on all evaluation metrics, with a 1.3\% M, 3.3\% C and 1\% RL performance increases. 
	
\eat{ and 2) from the quantitative results in Fig.\ref{Fig.example}, we can see that with S-LSTM our model can generate more precise and comprehensive sentences. This is because with S-LSTM layer, our model can approximate both prior distribution and posterior distribution as well as minimizing their distance in the training phrase.This enables the model to learn better parameters to approximate prior distribution for supporting efficient video captioning. Also, our stochastic layer models the uncertainties thus allowing describe video from different aspects.}  

\begin{table}[h]
	\centering
	\caption{Exploring MS-RNN. The top model uses only M-LSTM, while the bottom model integrates M-LSTM and S-LSTM. B, M, C, and RL are short for Blue, METEOR, CIDEr and ROUGE-L, respectively. All values are reported as percentage (\%).}
		\label{tab-setting}	
	\scalebox{1}{
	\begin{tabular}{c||c|c|c|c|c|c|c}
		\hline
		Model&B@1&B@2&B@3&B@4&M&C&RL\\
		\hline
		M& 82.7 & 71.8& 62.7 &52.2 &32.5 &71.5& 69.2\\
		M+S &\textbf{82.9} & \textbf{72.6}& \textbf{63.5} &\textbf{53.3}&\textbf{33.8}&\textbf{74.8}&\textbf{70.2}\\
		\hline
	\end{tabular}}

\end{table}

In Fig.\ref{five-rs}, we show some example sentences generated by our approach, with only M-LSTM and with both M-LSTM and S-LSTM, respectively. From Fig.\ref{five-rs}, we have the following observations:
\begin{itemize}
	\item Both M-LSTM and M-LSTM+S-LSTM are able to generate accurate descriptions for a video. In addition, the results generated by M-LSTM+S-LSTM are generally better than M-LSTM method, which is consistent with the results reported in Tab.~\ref{tab-setting}. 
	\item M-LSTM is deterministic and it can only generate one sentence, while M-LSTM+S-LSTM can produce different sentences.
	\item In general, M-LSTM+S-LSTM can provide more specific, comprehensive and accurate descriptions than M-LSTM. For example, the left top example, M-LSTM generates ``a women is playing a guitar'', while M-LSTM+S-LSTM provides ``a girl is singing'' and ``a women is playing with a guitar''. From the middle bottom, we can see that M-LSTM provides a wrong description ``cucumber'', while M-LSTM+S-LSTM generates ``vegetables'' and a set of verbs ``slicing, chopping and cutting''.
	\item Our MS-RNN model may produce duplicate and comprehensive results, which is consistent with the nature of video captioning.
	{
	\item The last column shows some wrong examples. For the right top example, both methods provide wrong descriptions, ``cutting a cucumber'' and ``slicing a carrot''. This is mainly because the MSVD dataset contains many videos about cooking and few videos about folding paper, which leads to an over-fitting problem, In addition, the right middle is also inaccurate. This is because both our models only take video appearance features as inputs and ignores the motion features. For the right bottom example, our model does not correctly identify the number of objects in some cases. } 
\end{itemize}

\begin{table}[t]
	\centering
	\caption{Comparing the quality of sentence generation on different video spatial representations on the MSVD dataset. (V), (G), and (R) stands for the VGGNet, GoogLeNet, and ResNe, respectively. This experiment is conducted on the MSVD dataset. All the values are reported as percentage (\%).}
	\scalebox{0.95}{
	\begin{tabular}{c||c|c|c|c|c|c}
		\hline
		Model&B@1&B@2&B@3&B@4&M&C\\
		\hline
		LSTM-E(V)\cite{CAP:EmbedPan2015Jointly}&74.9&60.9&50.6&40.2&29.5&-\\
		h-RNN(V)\cite{CAP:p_RNNYu_2016_CVPR}&77.3&64.5&54.6&44.3&31.1&62.1\\
		SA(G)\cite{CAP:softattYao2015Describing}&79.1&63.2&51.2&40.6&29.0&-\\
		MFA-LSTM(R)\cite{CAP:TM} &81.3& 69.8& 60.5& 50.4 & 32.2 & 69.8 \\		
		\hline
		MS-RNN(G)&80.3&68.4&58.7&48.0&31.0&66.6\\
		MS-RNN(R)&\textbf{82.9} & \textbf{72.6}& \textbf{63.5} &\textbf{53.3}&\textbf{33.8}&\textbf{74.8}\\
		\hline
	\end{tabular}}
	
	\label{tab-vf}
\end{table}
	
\subsection{Comparison Results on MSVD Dataset}
In this subsection, we conduct experiments to examine how different video representations work on video captioning, as well as comparing our model with existing approaches. In addition, all the approaches in this sub-experiments only take one type video representation extracting from VggNet (V), GoogleNet (G) or ResNet (R). We conduct our experiments on the MSVD dataset. 

Tab.~\ref{tab-vf} lists the experimental results. From Tab.\ref{tab-vf}, we have following observations: 
\begin{itemize}
	\item With only appearance features, our MS-RNN (R) model achieves the best performance on all evaluation metrics. Compared with the state-of-the-art method MFA-LSTM (R), our model achieves significantly better performance, with 1.6\%, 2.8\%, 3\%, 2.9\%, 1.6\% and 5\% increases on B@1, B@2, B@3, B@4, M and C, respectively. 
	\item For video captioning task, the RestNet-based video representation performs better than both VggNet-based and GoogleNet-based video features. Specifically, for our model RestNet feature performs better than GoogleNet features. For the whole experimental results, the approaches (SCN-LSTM and MFA-LSTM) with ResNet-based features performs better than the methods with GoogleNet or VggNet-based features. 
	\item  Compared with the methods using attention mechanisms, e.g., temporal attention \cite{CAP:softattYao2015Describing}, our MS-RNN (R) achieves even better results with 3.8\%, 9.4\%, 12.3\%, 12.7\% and  4.8\% increases on B@1, B@2, B@3, B@4 and M by using a simple mean pooling strategy. This indicates the advantages of our proposed MS-LSTM.
\end{itemize}

\begin{table}[t]
	\centering
	\caption{Performance comparison with methods using both appearance and motion video features. This experiment is conducted on the MSVD dataset.}
	\scalebox{0.95}{
	\begin{tabular}{c||c|c|c|c|c|c}
		\hline
		Model&B@1&B@2&B@3&B@4&M&C\\
		\hline
		LSTM-E(V+C)\cite{CAP:EmbedPan2015Jointly}&78.8&66.0&55.4&45.3&31.0&-\\
		SA(G+3D)\cite{CAP:softattYao2015Describing}&80.0&64.7&52.6&42.2&29.6&51.7\\
		h-RNN(V+C)\cite{CAP:p_RNNYu_2016_CVPR}&81.5&70.4&60.4&49.9&32.6&65.8\\
		MFA-LSTM(R+C)\cite{CAP:TM}&\textbf{82.9}&72.0&62.7&52.8&33.4&68.9\\
		SCN-LSTM(R+C) \cite{CAPs:SCN-LSTM}&-&-&-&51.1&33.5&\textbf{77.7}\\
		\hline
		\eat{
		MS-RNN(R)&\textbf{82.9} & \textbf{72.7}& \textbf{63.6} &\textbf{53.4}&\textbf{33.7}&74.4\\}
		MS-RNN(R)&\textbf{82.9} & \textbf{72.6}& \textbf{63.5} &\textbf{53.3}&\textbf{33.8}&\textbf{74.8}\\
		\hline
	\end{tabular}}
	
	\label{Tab.comparebest}
\end{table}

We also compare our methods with the others using multiple features. Specifically, in this subsection, we compare our model using only appearance features with six state-of-the-art methods: LSTM-E(V+C) \cite{CAP:EmbedPan2015Jointly}, SA(G+3DCNN) \cite{CAP:softattYao2015Describing}, HRNE-AT(G+C) \cite{CAP:HRNNPan2016Hierarchical}, h-RNN(V+C) \cite{CAP:p_RNNYu_2016_CVPR},	MFA-LSTM(R+C) \cite{CAP:TM} and SCN-LSTM(R+C) \cite{CAPs:SCN-LSTM}, which make use of both appearance and motion video features. Here, V and R are short for VggNet and ResNet, which are used to extract appearance features. 3D and C are short for 3DCNN and C3D, which are used to generate video motion features. 

The experimental results are shown in Tab.\ref{Tab.comparebest}. Although our model only uses appearance features, it performs better than existing methods on B@2 (72.6\%), B@3 (63.5\%), B@4 (53.3\%) and M (33.8\%), and achieves comparable results on B@1 (82.9\%) and C (74.8\%).

\begin{table}
	\centering
		\caption{Experiment results on the MSR-VTT dataset. SA-LSTM runs employ soft attention over the frame level features extracted from deep network, while MP-LSTM and our method utilize mean pooling over the frame level video features. }
	\begin{tabular}{c||c|c|c|c}
		\hline
		Model&B@4&M&C&RL\\
		\hline
		MP-LSTM(V)\cite{CAP:MP-LSTM}&34.8&24.8&-&-\\
		MP-LSTM(C)&35.4&24.8&-&-\\
		MP-LSTM(V+C)&35.8&25.3&-&-\\
		SA-LSTM(V)\cite{CAP:softattYao2015Describing}&35.6&25.4&-&-\\
		SA-LSTM(C)&36.1&25.7&-&-\\
		SA-LSTM(V+C)&36.6&25.9&-&-\\
		MFA-LSTM(R+C)\cite{CAP:TM}&39.2&\textbf{26.9}&\textbf{44.6}&\textbf{60.1}\\
		\hline
		\eat{
		MS-RNN (R)&38.8&25.3\\}
		MS-RNN(R)&\textbf{39.8}&26.1&40.9&59.3\\
		\hline
	\end{tabular}

	\label{Tab.MSRVTT}
\end{table}

\subsection{Comparison Results on MSR-VTT Dataset}
In this section, we compare our method with MP-LSTM \cite{CAP:MP-LSTM} and SA-LSTM \cite{CAP:softattYao2015Describing} on the MSR-VTT dataset. In addition, to obtain the appearance features, the MP-LSTM and our MS-RNN are based on the mean pooling strategy, while SA-LSTM is based on a soft attention mechanism. In theory, soft attention is more complex than mean pooling, but usually provides better visual features. The experimental results are shown in Tab.\ref{Tab.MSRVTT} and we have the following observations:
\begin{itemize}
	\item MS-RNN gains a promising performance with 39.8\% B@4, 26.1\% M, 40.9\% C and 59.3\% RL  \eat{38.8\% B@4, 25.3\% M, 38.8\% C and 58.2\% RL} on the MSR-VTT dataset.
	\item Overall with same visual input (VGG-19, VGG-19+C3D, or C3D), SA-LSTM performs better than MP-LSTM. However, SA is based on the soft attention. In other words, in theory SA-LSTM takes better visual features as inputs. Compared with MP-LSTM, our MS-RNN (R) outperforms MP-LSTM (VGG-19+C3D) with 4\% B@4 and 0.8\% M increases. Compared with SA-LSTM, our MS-RNN (R) outperforms SA-LSTM(VGG-19+C3D) with 3.2\% B@4.
	{
	Compared with MS-RNN(R+C), our model achieves comparable results on B@4, M and RL by using single feature (R).}
	
\end{itemize}

\section{Conclusions and Future Work}
In this paper, we propose a Multimodal Stochastic Recurrent Neural Network (MS-RNN) framework for video captioning. This work has shown how to extend the modeling capabilities of RNN by approximating both prior distribution and true posterior distribution with a nonlinear latent layer (S-LSTM). In addition, MS-RNN achieves the state-of-the-art performance with only mean video appearance features and is comparable with the counterparts, which take both video appearance and motion features. Last but not least, the proposed model can be applied to a wide range of video analysis applications. 

In the future, we will integrate the state-of-the-art attention mechanism~\cite{CAP:softattYao2015Describing} with our model to further improve the video captioning performance. Moreover, the motion feature will be considered.

\ifCLASSOPTIONcaptionsoff
  \newpage
\fi



\begin{thebibliography}{10}
	\providecommand{\url}[1]{#1}
	\csname url@samestyle\endcsname
	\providecommand{\newblock}{\relax}
	\providecommand{\bibinfo}[2]{#2}
	\providecommand{\BIBentrySTDinterwordspacing}{\spaceskip=0pt\relax}
	\providecommand{\BIBentryALTinterwordstretchfactor}{4}
	\providecommand{\BIBentryALTinterwordspacing}{\spaceskip=\fontdimen2\font plus
		\BIBentryALTinterwordstretchfactor\fontdimen3\font minus
		\fontdimen4\font\relax}
	\providecommand{\BIBforeignlanguage}[2]{{%
			\expandafter\ifx\csname l@#1\endcsname\relax
			\typeout{** WARNING: IEEEtran.bst: No hyphenation pattern has been}%
			\typeout{** loaded for the language `#1'. Using the pattern for}%
			\typeout{** the default language instead.}%
			\else
			\language=\csname l@#1\endcsname
			\fi
			#2}}
	\providecommand{\BIBdecl}{\relax}
	\BIBdecl
	
	\bibitem{CAP:Kojima2002Natural}
	A.~Kojima, T.~Tamura, and K.~Fukunaga, ``Natural language description of human
	activities from video images based on concept hierarchy of actions,''
	\emph{International Journal of Computer Vision}, vol.~50, no.~2, pp.
	171--184, 2002.
	
	\bibitem{Tra:Lee2008SAVE}
	M.~W. Lee, A.~Hakeem, N.~Haering, and S.~Zhu, ``{SAVE:} {A} framework for
	semantic annotation of visual events,'' in \emph{Computer Vision and Patter
		Recognition}, 2008, pp. 1--8.
	
	\bibitem{Tra:Khan2011Human}
	M.~U.~G. Khan, L.~Zhang, and Y.~Gotoh, ``Human focused video description,'' in
	\emph{International Conference on Computer Vision}, 2011, pp. 1480--1487.
	
	\bibitem{Tra:Hanckmann2012Automated}
	P.~Hanckmann, K.~Schutte, and G.~J. Burghouts, ``Automated textual descriptions
	for a wide range of video events with 48 human actions,'' in \emph{ECCV},
	2012, pp. 372--380.
	
	\bibitem{Net:VGGSimonyan2014Very}
	K.~Simonyan and A.~Zisserman, ``Very deep convolutional networks for
	large-scale image recognition,'' in \emph{ICLR}, 2014.
	
	\bibitem{Net:ResNetHe_2016_CVPR}
	K.~He, X.~Zhang, S.~Ren, and J.~Sun, ``Deep residual learning for image
	recognition,'' in \emph{Computer Vision and Patter Recognition}, 2016, pp.
	770--778.
	
	\bibitem{RNNs:LSTMsep}
	S.Hochreiter and J.Schmidhuber, ``Long short-term memory,'' \emph{Neural
		Computation}, vol.~9, no.~8, pp. 1735--1780, 1997.
	
	\bibitem{RNNs:GRUJun}
	J.~Chung, {\c{C}}.~G{\"{u}}l{\c{c}}ehre, K.~Cho, and Y.~Bengio, ``Empirical
	evaluation of gated recurrent neural networks on sequence modeling,''
	\emph{CoRR}, vol. abs/1412.3555, 2014.
	
	\bibitem{CAPs:seq2seq}
	S.~Venugopalan, M.~Rohrbach, J.~Donahue, R.~J. Mooney, T.~Darrell, and
	K.~Saenko, ``Sequence to sequence - video to text,'' in \emph{International
		Conference on Computer Vision}, 2015, pp. 4534--4542.
	
	\bibitem{CAP:EmbedPan2015Jointly}
	Y.~Pan, T.~Mei, T.~Yao, H.~Li, and Y.~Rui, ``Jointly modeling embedding and
	translation to bridge video and language,'' in \emph{Computer Vision and
		Patter Recognition}, 2016, pp. 4594--4602.
	
	\bibitem{CAP:softattYao2015Describing}
	L.~Yao, A.~Torabi, K.~Cho, N.~Ballas, C.~J. Pal, H.~Larochelle, and A.~C.
	Courville, ``Describing videos by exploiting temporal structure,'' in
	\emph{International Conference on Computer Vision}, 2015, pp. 4507--4515.
	
	\bibitem{CAP:Guo2016Attention}
	Z.~Guo, L.~Gao, J.~Song, X.~Xu, J.~Shao, and H.~T. Shen, ``Attention-based
	{LSTM} with semantic consistency for videos captioning,'' in \emph{ACM MM},
	2016, pp. 357--361.
	
	\bibitem{CAP:TM}
	X.~Long, C.~Gan, and G.~de~Melo, ``Video captioning with multi-faceted
	attention,'' \emph{CoRR}, vol. abs/1612.00234, 2016.
	
	\bibitem{ImCap:Attri}
	Q.~You, H.~Jin, Z.~Wang, C.~Fang, and J.~Luo, ``Image captioning with semantic
	attention,'' in \emph{Computer Vision and Patter Recognition}, 2016, pp.
	4651--4659.
	
	\bibitem{CAP:BAttri}
	T.~Yao, Y.~Pan, Y.~Li, Z.~Qiu, and T.~Mei, ``Boosting image captioning with
	attributes,'' \emph{CoRR}, vol. abs/1611.01646, 2016.
	
	\bibitem{CAP:HRNNPan2016Hierarchical}
	P.~Pan, Z.~Xu, Y.~Yang, F.~Wu, and Y.~Zhuang, ``Hierarchical recurrent neural
	encoder for video representation with application to captioning,'' in
	\emph{Computer Vision and Patter Recognition}, 2016, pp. 1029--1038.
	
	\bibitem{CAP:p_RNNYu_2016_CVPR}
	H.~Yu, J.~Wang, Z.~Huang, Y.~Yang, and W.~Xu, ``Video paragraph captioning
	using hierarchical recurrent neural networks,'' in \emph{Computer Vision and
		Patter Recognition}, 2016, pp. 4584--4593.
	
	\bibitem{VAEs:Bayes}
	D.~P. Kingma and M.~Welling, ``Auto-encoding variational bayes,'' in
	\emph{ICLR}, 2014.
	
	\bibitem{RNNs:simpleRNN_2}
	J.~L. Elman, ``Finding structure in time,'' \emph{Cognitive Science}, vol.~14,
	no.~2, pp. 179--211, 1990.
	
	\bibitem{RNNs:simpleRNN_1}
	M.~I. Jordan, ``Serial order: A parallel distributed processing approach.''
	vol. 121, p.~64, 1986.
	
	\bibitem{RNNs:RNNLM_Tomas}
	T.~Mikolov, M.~Karafi{\'{a}}t, L.~Burget, J.~Cernock{\'{y}}, and S.~Khudanpur,
	``Recurrent neural network based language model,'' in \emph{INTERSPEECH,},
	2010, pp. 1045--1048.
	
	\bibitem{Tra:ImCapMRF}
	A.~Farhadi, S.~M.~M. Hejrati, M.~A. Sadeghi, P.~Young, C.~Rashtchian,
	J.~Hockenmaier, and D.~A. Forsyth, ``Every picture tells a story: Generating
	sentences from images,'' in \emph{ECCV}, 2010, pp. 15--29.
	
	\bibitem{Tra:CRF}
	M.~Rohrbach, W.~Qiu, I.~Titov, S.~Thater, M.~Pinkal, and B.~Schiele,
	``Translating video content to natural language descriptions,'' in
	\emph{International Conference on Computer Vision}, 2013, pp. 433--440.
	
	\bibitem{Net:GNetSzegedy_2015_CVPR}
	C.~Szegedy, W.~Liu, Y.~Jia, P.~Sermanet, S.~E. Reed, D.~Anguelov, D.~Erhan,
	V.~Vanhoucke, and A.~Rabinovich, ``Going deeper with convolutions,'' in
	\emph{Computer Vision and Patter Recognition}, 2015, pp. 1--9.
	
	\bibitem{CAP:MP-LSTM}
	S.~Venugopalan, H.~Xu, J.~Donahue, M.~Rohrbach, R.~J. Mooney, and K.~Saenko,
	``Translating videos to natural language using deep recurrent neural
	networks,'' in \emph{NAACL HLT}, 2015, pp. 1494--1504.
	
	\bibitem{gan2017stylenet}
	C.~Gan, Z.~Gan, X.~He, J.~Gao, and L.~Deng, ``Stylenet: Generating attractive
	visual captions with styles,'' in \emph{CVPR}, 2017, pp. 3137--3146.
	
	\bibitem{liang2017recurrent}
	X.~Liang, Z.~Hu, H.~Zhang, C.~Gan, and E.~P. Xing, ``Recurrent topic-transition
	gan for visual paragraph generation,'' \emph{arXiv preprint
		arXiv:1703.07022}, 2017.
	
	\bibitem{li2017temporal}
	F.~Li, C.~Gan, X.~Liu, Y.~Bian, X.~Long, Y.~Li, Z.~Li, J.~Zhou, and S.~Wen,
	``Temporal modeling approaches for large-scale youtube-8m video
	understanding,'' \emph{arXiv preprint arXiv:1707.04555}, 2017.
	
	\bibitem{ImCap:Diver}
	Z.~Wang, F.~Wu, W.~Lu, J.~Xiao, X.~Li, Z.~Zhang, and Y.~Zhuang, ``Diverse image
	captioning via grouptalk,'' in \emph{IJCAI}, 2016, pp. 2957--2964.
	
	\bibitem{KendallGal2017Uncertainties}
	A.~Kendall and Y.~Gal, ``What uncertainties do we need in bayesian deep
	learning for computer vision?'' vol. abs/1703.04977, 2017.
	
	\bibitem{LiGal2017Alpha}
	Y.~Li and Y.~Gal, ``Dropout inference in bayesian neural networks with
	alpha-divergences,'' vol. abs/1703.02914, 2017.
	
	\bibitem{Gal2016Improving}
	Y.~Gal, R.~McAllister, and C.~E. Rasmussen, ``Improving {PILCO} with {B}ayesian
	neural network dynamics models,'' in \emph{Data-Efficient Machine Learning
		workshop, ICML}, 2016.
	
	\bibitem{Uusitalo201524}
	L.~Uusitalo, A.~Lehikoinen, I.~Helle, and K.~Myrberg, ``An overview of methods
	to evaluate uncertainty of deterministic models in decision support,''
	\emph{Environmental Modelling and Software}, vol.~63, pp. 24 -- 31, 2015.
	
	\bibitem{CVAE:Kihyuk}
	K.~Sohn, H.~Lee, and X.~Yan, ``Learning structured output representation using
	deep conditional generative models,'' in \emph{NIPS}, 2015, pp. 3483--3491.
	
	\bibitem{VAE:deepkf}
	R.~G. Krishnan, U.~Shalit, and D.~Sontag, ``Deep kalman filters,'' \emph{CoRR},
	vol. abs/1511.05121, 2015.
	
	\bibitem{VAE:HVRNNIulian}
	I.~V. Serban, A.~Sordoni, R.~Lowe, L.~Charlin, J.~Pineau, A.~C. Courville, and
	Y.~Bengio, ``A hierarchical latent variable encoder-decoder model for
	generating dialogues,'' in \emph{AAAI}, 2017, pp. 3295--3301.
	
	\bibitem{VAE:SRNNMarco}
	M.~Fraccaro, S.~K. S{\o}nderby, U.~Paquet, and O.~Winther, ``Sequential neural
	models with stochastic layers,'' in \emph{NIPS}, 2016, pp. 2199--2207.
	
	\bibitem{VAE:VRNNJunyoung}
	J.~Chung, K.~Kastner, L.~Dinh, K.~Goel, A.~C. Courville, and Y.~Bengio, ``A
	recurrent latent variable model for sequential data,'' in \emph{NIPS}, 2015,
	pp. 2980--2988.
	
	\bibitem{Nets:faster-rcnn}
	S.~Ren, K.~He, R.~Girshick, and J.~Sun, ``Faster {R-CNN}: Towards real-time
	object detection with region proposal networks,'' in \emph{NIPS}, 2015.
	
	\bibitem{ImCap:LearingRNN}
	X.~Chen and C.~L. Zitnick, ``Learning a recurrent visual representation for
	image caption generation,'' \emph{CoRR}, vol. abs/1411.5654, 2014.
	
	\bibitem{CAPs:From_captions}
	H.~Fang, S.~Gupta, F.~N. Iandola, R.~K. Srivastava, L.~Deng, P.~Doll{\'{a}}r,
	J.~Gao, X.~He, M.~Mitchell, J.~C. Platt, C.~L. Zitnick, and G.~Zweig, ``From
	captions to visual concepts and back,'' in \emph{Computer Vision and Patter
		Recognition}, 2015, pp. 1473--1482.
	
	\bibitem{OPT:ADADELTA}
	M.~D. Zeiler, ``{ADADELTA:} an adaptive learning rate method,'' \emph{CoRR},
	vol. abs/1212.5701, 2012.
	
	\bibitem{Dataset:msvd}
	D.~Chen and W.~B. Dolan, ``Collecting highly parallel data for paraphrase
	evaluation,'' in \emph{ACL HLT}, 2011, pp. 190--200.
	
	\bibitem{Dataset:msr-vtt}
	J.~Xu, T.~Mei, T.~Yao, and Y.~Rui, ``{MSR-VTT:} {A} large video description
	dataset for bridging video and language,'' in \emph{Computer Vision and
		Patter Recognition}, 2016, pp. 5288--5296.
	
	\bibitem{EVA:BLUE}
	K.~Papineni, S.~Roukos, T.~Ward, and W.~Zhu, ``Bleu: a method for automatic
	evaluation of machine translation,'' in \emph{ACL}, 2002, pp. 311--318.
	
	\bibitem{EVA:METEOR}
	M.~J. Denkowski and A.~Lavie, ``Meteor universal: Language specific translation
	evaluation for any target language,'' in \emph{The Workshop on Statistical
		Machine Translation,}, 2014, pp. 376--380.
	
	\bibitem{EVA:CIDEr}
	R.~Vedantam, C.~L. Zitnick, and D.~Parikh, ``Cider: Consensus-based image
	description evaluation,'' in \emph{Computer Vision and Patter Recognition},
	2015, pp. 4566--4575.
	
	\bibitem{EVA:ROUGE}
	C.~Flick, ``Rouge: A package for automatic evaluation of summaries,'' in
	\emph{The Workshop on Text Summarization Branches Out}, 2004, p.~10.
	
	\bibitem{EVA:MCOCO}
	X.~Chen, H.~Fang, T.~Y. Lin, R.~Vedantam, S.~Gupta, P.~Dollar, and C.~L.
	Zitnick, ``Microsoft coco captions: Data collection and evaluation server,''
	\emph{Computer Science}, 2015.
	
	\bibitem{method:BeamSearch}
	D.~Furcy and S.~Koenig, ``Limited discrepancy beam search,'' in \emph{IJCAI},
	2005, pp. 125--131.
	
	\bibitem{CAPs:SCN-LSTM}
	Z.~Gan, C.~Gan, X.~He, Y.~Pu, K.~Tran, J.~Gao, L.~Carin, and L.~Deng,
	``Semantic compositional networks for visual captioning,'' \emph{CoRR}, vol.
	abs/1611.08002, 2016.
	
\end{thebibliography}
\end{document}